\documentclass[letterpaper, 10 pt, conference]{ieeeconf}  

\IEEEoverridecommandlockouts                              

\usepackage{algorithm}
\usepackage{algpseudocode}

\usepackage{listings}
\usepackage{xcolor}  

\usepackage{graphics}
\usepackage{listings}
\usepackage{cite}
\usepackage{graphicx}
\usepackage{hyperref}
\usepackage{amsmath}
\usepackage[position=bottom]{subfig}
\graphicspath{ {images/} }

\usepackage{tcolorbox}
\usepackage{amsmath}

\newtcolorbox{promptbox}{colframe=black, colback=white, boxrule=0.5pt, left=5pt, right=5pt, top=3pt, bottom=3pt}

\lstdefinestyle{prolog}{
    language=Prolog,
    basicstyle=\ttfamily\footnotesize,
    keywordstyle=\color{blue},
    commentstyle=\color{green},
    stringstyle=\color{red},
    numbers=left,
    numberstyle=\tiny,
    stepnumber=1,
    numbersep=5pt,
    breaklines=true,
    frame=single,
}

\title{\LARGE \bf
RecipeMasterLLM: Revisiting RoboEarth in the Era of Large Language Models
}

\author{Asil Kaan Bozcuo\u{g}lu* and Ziyuan Liu*
\thanks{*A. Bozcuo\u{g}lu and Z. Liu are with Huawei Heisenberg Research Center (Munich),
        Riesstr. 25, 80992 Munich, Germany
        {\tt\small {asil.kaan.bozcuoglu, ziyuan.liu1}@huawei.com}}%
}

\begin{document}

\maketitle
\thispagestyle{empty}
\pagestyle{empty}

\begin{abstract}
RoboEarth~\cite{tenorth13tase} was a pioneering initiative in cloud robotics, establishing a foundational framework for robots to share and exchange knowledge about actions, objects, and environments through a standardized knowledge graph. Initially, this knowledge was predominantly hand-crafted by engineers using RDF triples within OWL Ontologies~\cite{antoniou2009web}, with updates, such as changes in an object's pose, being asserted by the robot's control and perception routines. However, with the advent and rapid development of Large Language Models (LLMs), we believe that the process of knowledge acquisition can be significantly automated. To this end, we propose \textit{RecipeMasterLLM}, a high-level planner, that generates OWL action ontologies based on a standardized knowledge graph in response to user prompts. This architecture leverages a fine-tuned LLM specifically trained to understand and produce action descriptions consistent with the RoboEarth standardized knowledge graph. Moreover, during the Retrieval-Augmented Generation (RAG) phase, environmental knowledge is supplied to the LLM to enhance its contextual understanding and improve the accuracy of the generated action descriptions.

\end{abstract}

\section{INTRODUCTION}

Large Language Models (LLMs) are powerful tools, enabling humans to access and process vast amounts of knowledge with ease. For robots to achieve similar capabilities, it is crucial to establish a shared foundation of symbolic groundings between LLMs and robotic control systems so that a shared semantics communication layer is achieved. This requires aligning their knowledge representations with standardized toolkits, such as knowledge graphs, that integrate seamlessly with robot control executives. 
Aligned symbolic groundings enable LLMs to represent and communicate knowledge seamlessly within robotic systems.

In the early 2010s, a notable European project called RoboEarth~\cite{tenorth13tase} emerged. In this project, the researchers developed a comprehensive robotic knowledge representation and reasoning framework, along with a standardized base knowledge graph, enabling robots to utilize existing knowledge about actions, objects, and environments, and to share this knowledge with one another. RoboEarth was, in essence, envisioned as the "World Wide Web" for robots at the time. These advancements were later integrated into an open-source software framework, compatible with Robot Operating System (ROS), known as KnowRob~\cite{beetz18knowrob}.

A key use case~\cite{bozcuoglu2018exchange, tenorth13tase} involves robots downloading generic action descriptions from RoboEarth servers and adapting these descriptions to their specific environments using existing environmental knowledge. However, a significant limitation at the time was that most of the knowledge in these descriptions had to be crafted manually. Although modeling generic domain knowledge, such as common sense knowledge for household environments, as a foundational knowledge graph is feasible as a one-time effort, this approach is not scalable, particularly for action descriptions. Manually modeling the vast array of actions required to achieve numerous goals in unstructured environments quickly becomes impractical.

\begin{figure}[!t]
    \centering
    \includegraphics[width=\linewidth]{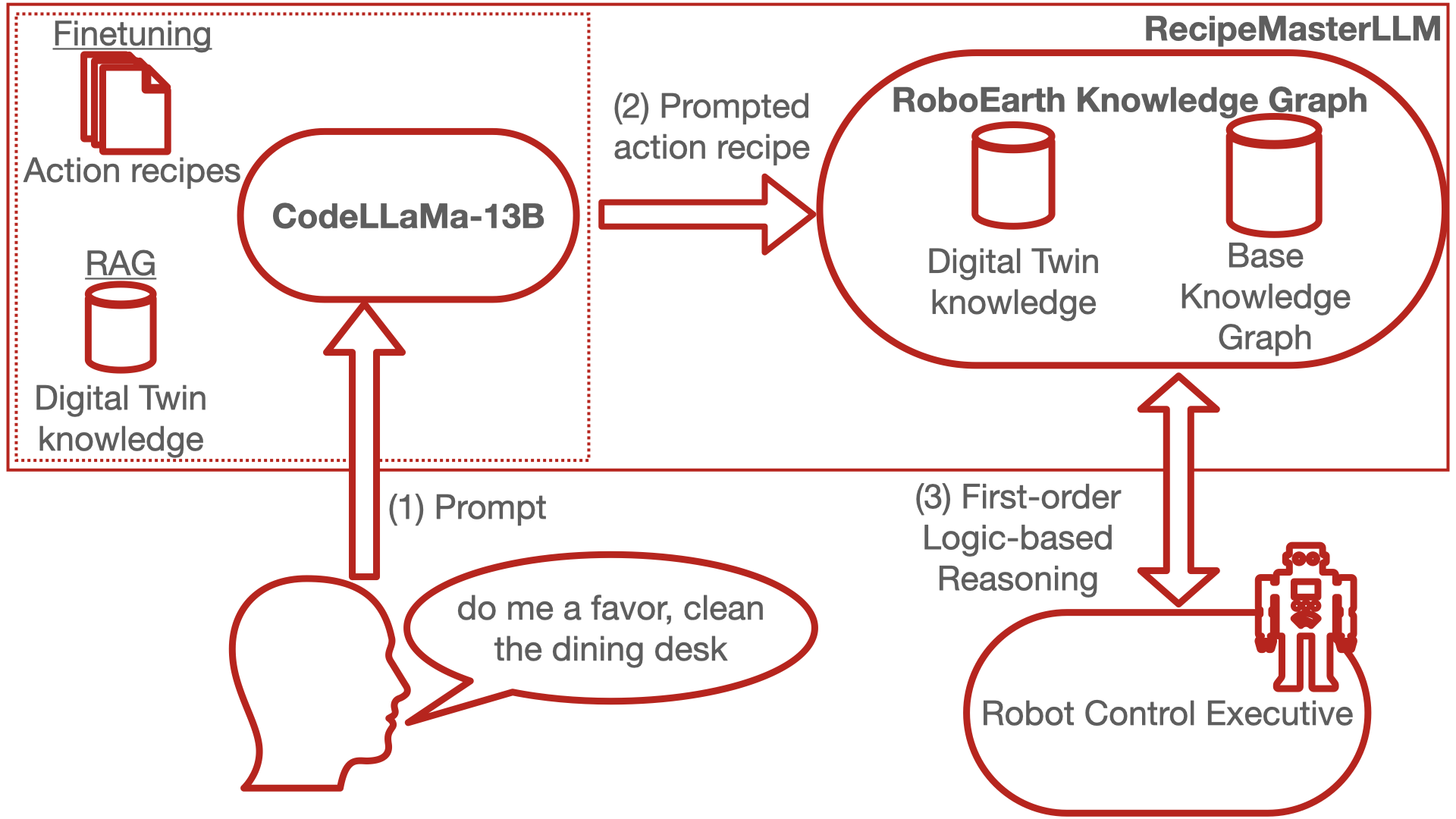}
    \caption{\small High-level concept diagram of the proposed system: The user provides a high-level goal to be delegated to the robot. This goal is processed by a fine-tuned large language model (LLM), which references the environment's semantic map (digital twin knowledge). The resulting action plan is then integrated into the RoboEarth inference system.}
    \label{fig:idea}
\end{figure}

LLMs, on the other hand, present a promising solution for generating and curating the vast amounts of 
on-demand knowledge required by robots operating in unstructured environments. These models are typically trained and fine-tuned
on extensive datasets, making them a rich source of knowledge. The shift towards automated knowledge generation through LLMs has
the potential to significantly enhance the adaptability and functionality of robotic systems, thereby advancing the capabilities 
of cloud robotics in complex, real-world scenarios. A significant challenge, as aforementioned, in integrating this curated knowledge lies in grounding it in robot control, as LLMs typically provide
their outputs in natural language. We believe that the foundational work of RoboEarth offers a robust framework for addressing
this issue. By utilizing the terminologies and constraints defined in a standardized knowledge graph, RoboEarth's framework can
enable LLMs to express the required knowledge in a format that is more compatible with robot control systems.


In this paper, we introduce \textit{RecipeMasterLLM}, a high-level planner designed to inject structured, high-level plans into a full-scale robotic knowledge graph and reasoning framework. This is achieved by fine-tuning a \textbf{small-scale, open-source LLM}, \textbf{enabling seamless deployment in cloud robotics environments without reliance on large proprietary LLM providers or large GPU clusters}. Through this approach, LLMs serve as an additional knowledge source, capable of representing information using terminologies (TBOX) and assertions (ABOX) derived from fine-tuning and Retrieval-Augmented Generation (RAG). This empowers LLMs to perform long-horizon task planning—\textit{action recipes} in RoboEarth—based on user prompts, while ensuring symbolic alignment with the RoboEarth Knowledge Graph (RKG) (Figure~\ref{fig:idea}). As a result, \textbf{LLMs, engineers, and domain experts collaboratively contribute to the RoboEarth knowledge graph}, enriching its knowledge base and reasoning capabilities.  

Building on this foundation, we present the following contributions: (1) a pipeline-based architecture that enables robots to plan and act based on user prompts, and (2) a fine-tuned derivative of CodeLLaMa~\cite{roziere2023code}, specifically designed to generate action descriptions, or \textit{action recipes} as defined in RoboEarth. This model is further enhanced through exposure to environmental knowledge—referred to as \textit{digital twin knowledge}—during the Retrieval-Augmented Generation (RAG) phase.

\section{STATE OF THE ART}



Han et al.~\cite{han2024interpretinteractivepredicatelearning} introduce InterPreT, an LLM-powered framework that enables robots to learn symbolic predicates and operators from human language feedback during interaction, facilitating long-horizon planning and generalizing effectively to complex tasks in both simulated and real-world environments.

Similarly, Chen et al.~\cite{chen2024languageaugmentedsymbolicplanneropenworld} propose the Language-Augmented Symbolic Planner (LASP), which integrates pre-trained LLMs with symbolic planners to address the limitations of planning in open-world environments with incomplete domain knowledge. LASP effectively diagnoses execution errors and incrementally builds its knowledge base, allowing robotic agents to tackle complex, long-horizon tasks despite multiple knowledge gaps.

Ahn et al.~\cite{ahn2022can} highlight a key challenge in leveraging LLMs for robotic planning: while LLMs encode vast amounts of semantic knowledge, they lack real-world grounding, which limits their direct applicability to robotic tasks. To address this, they propose using pretrained skills as a grounding mechanism, allowing robots to act as the "hands and eyes" of the language model while leveraging its high-level reasoning capabilities. Their approach combines low-level robotic skills with LLM-driven planning, demonstrating effective execution of complex, temporally extended instructions in real-world tasks.

In \cite{gao2024dag}, Gao et al. introduce DAG-Plan, a task planning framework designed specifically for dual-arm robots. Leveraging large language models (LLMs) to decompose complex tasks into a directed acyclic graph (DAG) of actionable sub-tasks, DAG-Plan dynamically allocates tasks to each arm based on real-time observations, enabling parallel and adaptive execution. Their evaluation on the Dual-Arm Kitchen Benchmark demonstrates nearly 50\% higher efficiency compared to single-arm baselines and double the success rate over dual-arm task planning baselines.

Wang et al.~\cite{wang2024llmbasedrobottaskplanning} introduce a task planning method using a constrained LLM prompt scheme to generate executable action sequences from natural language commands. They also propose an exception handling module to address LLM hallucinations, ensuring the generated plans are valid in real-world environments.

Sun et al.~\cite{sun2024prompt} present a framework that combines adversarial imitation learning with LLMs to enable humanoid robots to learn reusable skills with a single policy. This approach, enhanced by vector quantization and general reward functions, allows robots to perform zero-shot tasks through LLM-guided prompts and adapt efficiently to complex motion tasks without the need for multiple policies or additional guiding mechanisms.

Lastly, Kannan et al.~\cite{kannan2024smartllmsmartmultiagentrobot} propose SMART-LLM, a framework for multi-robot task planning that leverages LLMs to translate high-level task instructions into actionable plans through task decomposition, coalition formation, and task allocation. Their evaluations in both simulation and real-world scenarios demonstrate SMART-LLM’s effectiveness, validated by a benchmark dataset covering tasks of varying complexity.

\begin{figure*}[!t]
    \centering
    \includegraphics[width=\linewidth]{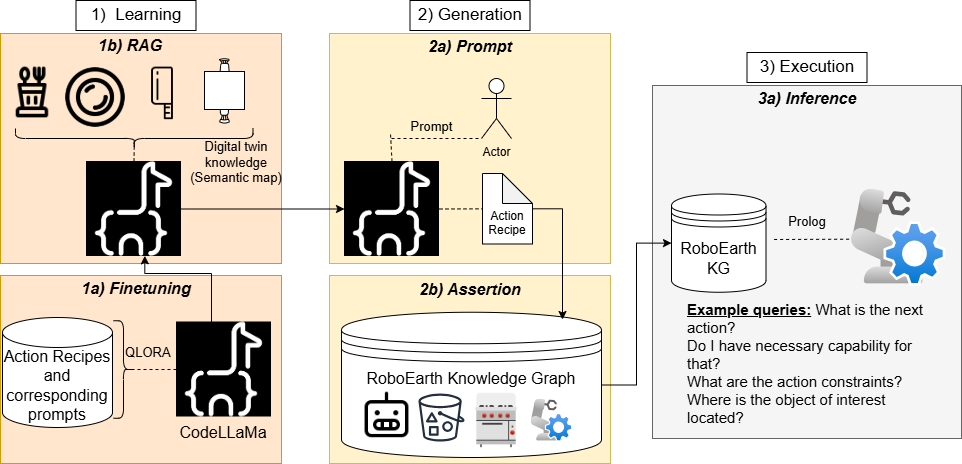}
    \caption{\small End-to-end pipeline during robot execution. Before execution, the semantic map of the environment is integrated into the RoboEarth Inference System. The process is triggered by a user prompt, which generates and asserts an \textit{action recipe}. The robot then infers the necessary actions and retrieves relevant knowledge based on the provided prompt.}
    \label{fig:architecture}
\end{figure*}
\section{SYSTEM}

The RoboEarth knowledge graph (RKG) serves as a tool set for representing, organizing, and inferring the knowledge required for robotic tasks. A central component of this framework is the concept of \textit{action recipes}, which provide robots with structured, task-specific instructions to execute user-delegated manipulation and navigation tasks. These recipes establish a foundational understanding that can evolve through learning, ensuring both reliable and consistent robot behavior while supporting autonomy.

In Figure~\ref{fig:architecture}, we present the different stages of \textit{RecipeMasterLLM}. In \textit{Stage 1A}, we fine-tune CodeLLaMa using our \textit{action recipe} dataset, which includes examples from household environments. This fine-tuned CodeLLaMa then indexes the digital twin knowledge in \textit{Stage 1B}, becoming environment-aware and ready to respond to user prompts. In the subsequent steps (Stages 2A-B), CodeLLaMa accepts a prompt, generates the corresponding \textit{action recipe}, and asserts it to the RoboEarth Knowledge Graph (RKG). The prompts are converted into executable robotic plans, enhancing both the adaptability and scalability of task planning in dynamic environments. During runtime (Stage 3A), the robot control executive (RCE) queries the RKG to determine how to execute the asserted \textit{action recipe}. The RKG's responses are based on both the asserted \textit{action recipe} and the base knowledge graph.

\subsection{Large Language Model (CodeLLaMa)} 

To implement the proposed system on private cloud or compute infrastructure with limited resources, we selected the open-source, small-scale LLM CodeLLaMa 13B~\cite{roziere2023code}, chosen for its pre-training on extensive code datasets. We anticipated that this would facilitate easier generalization of \textit{action recipe} generation compared to other open-source LLMs like LLaMa3.1-8B. However, initial tests revealed insufficient knowledge in RoboEarth or KnowRob. To address this, we developed a two-stage methodology.


In the first stage, we employ the VRAM-efficient QLORA fine-tuning technique~\cite{dettmers2024qlora}, which significantly reduces memory usage through quantization. This method compresses the model’s weights into lower precision formats (e.g., 4-bit), thereby enabling fine-tuning on large models without the need for extensive hardware resources. This approach facilitates efficient fine-tuning of CodeLLaMa on our action recipe dataset, optimizing both performance and resource consumption.


For fine-tuning, we curated an \textit{action recipe} dataset in the OpenAssistant Guanaco format~\cite{kpf2023openassistantconversationsdemocratizing}. This dataset includes \textit{action recipes} from RoboEarth, as well as auto-generated recipes created using GPT-4 through prompt engineering. In a way, this process acts as \textbf{knowledge distillation} from a large teacher LLM to a smaller student LLM. Each recipe is thoroughly reviewed before being included in the dataset. The final dataset consists of 350 distinct action recipes, along with the corresponding user prompts used for their generation. An illustration of the data instance with the corresponding interpretation of its \textit{action recipe} in natural language is as follows:

\begin{lstlisting}
{
  "user_prompt": "pick up the cup from 
  the table and place it on the sink?",
  "action_recipe": {
    "approach to table_1",
    "grasp cup_1",
    "move towards sink_1",
    "place cup_1 on sink_1"
  }
}
\end{lstlisting}

In the second stage, we incorporate a Retrieval-Augmented Generation (RAG) approach using LlamaIndex~\footnote{https://www.llamaindex.ai/} and the \texttt{bge-small-en-v1.5} embedding model~\footnote{https://huggingface.co/BAAI/bge-small-en-v1.5} to enhance the model’s ability to generate \textit{action recipes} based on the most current digital twin knowledge. The RAG method combines the strengths of large language models with external knowledge sources, ensuring that the generated \textit{action recipes} are enriched with relevant, up-to-date knowledge. Furthermore, this approach offers significant advantages in terms of speed over fine-tuning, allowing the model to quickly adapt to new environments. LlamaIndex serves as an interface that connects the LLM to both structured and unstructured data sources, enabling it to retrieve semantic information from the environment’s knowledge base. The \texttt{bge-small-en-v1.5} embedding model transforms textual data into a vector space, facilitating the efficient retrieval of the most contextually appropriate information. This setup enhances the system’s accuracy and contextual understanding, particularly when handling complex or highly abstract user prompts.

\subsection{RoboEarth: Knowledge Graph and Inference System}

The RKG is a knowledge processing system designed to integrate knowledge representation and reasoning with methods for knowledge acquisition and grounding in physical systems. Serving as a unified semantic framework, it allows the integration of diverse information sources and is built on the Web Ontology Language (OWL)~\footnote{https://www.w3.org/TR/owl-ref/}, following the World Wide Web Consortium's (W3C) standard for data representation known as the Resource Description Framework (RDF). RDF structures data as a directed graph composed of triple statements, which can be dynamically asserted or updated during runtime through queries.

The ability to represent, acquire, and reason with knowledge is critical for robots to perform complex, context-aware tasks autonomously. the RKG facilitates this by enabling robots to model environments, infer task sequences, and recognize object affordances. By integrating both symbolic knowledge and real-time sensory input, the RKG encodes this knowledge in OWL, following the RDF model, and supports rich semantic definitions. This approach enables the creation of a unified ontology of actions, objects, and their interrelations within robotic environments, ensuring that robots have a structured, adaptable framework for understanding their world.

A key strength of the RKG lies in its capacity to dynamically expand a robot’s knowledge base through interaction and learning. The knowledge is structured as a graph of RDF triples, making it flexible to updates and expansions based on the robot’s experiences and input from external sources. Inference capabilities, powered by SWI-Prolog, allow robots to reason logically over this data, formulate complex queries, and deduce actionable steps from abstract knowledge. These features are crucial for enabling robots to adapt to new environments, perform task reasoning, and execute tasks autonomously with a high degree of flexibility.

The \textit{action recipes} represent a fundamental component of this framework which specify task-specific sequences of robotic actions, providing robots with structured, step-by-step guidance for executing complex tasks such as object manipulation, navigation, and environmental interaction. Each recipe includes details on the necessary actions, objects, and contexts required to complete the task. When the RKG was first released, these action recipes were crafted by human experts using a dedicated editor, ensuring that robots began with a foundational understanding of tasks grounded in human logic and expectations. Over time, this understanding could be extended and refined through the robot’s own observations and learning, enabling dynamic adaptation to new tasks and environments. With the introduction of our LLM-enabled pipeline, the generation of action recipes is now automated, eliminating the need for manual encoding and significantly improving scalability and efficiency. This automation not only accelerates the creation of action recipes but also allows for their contextual adaptation to the robot’s environment in real time. The use of action recipes, whether human-encoded or generated automatically, ensures consistency and reliability in robotic behavior, providing a stable basis for further autonomous adaptation.

To query the RKG, we use SWI-Prolog, a logic programming language that supports backtracking, where variables can be bound (assigned a specific value) or unbound (undefined), with the system attempting to bind unbound variables through pattern matching and logical inference. SWI-Prolog also includes extensive libraries for applications like constraint logic, natural language processing, and knowledge representation. The robot control executive (RCE) formulates queries in SWI-Prolog, enabling effective interaction with and reasoning over the knowledge graph.

\subsection{Robot Control Executive}

The RCE  is responsible for orchestrating the execution of the proposed architecture from start to finish. Its primary function is to monitor for new actions to be performed. Upon detecting a new task, it queries the knowledge graph to retrieve the relevant parameters and contextual knowledge necessary for executing the identified actions (see Algorithm~\ref{alg:mainloop}).

\begin{algorithm}
\caption{The Main Loop of the Robot Control Executive (RCE)}\label{alg:mainloop}
\begin{algorithmic}[1]
\Procedure{Main Loop}{}
    \State action = first\_action
    \While{action is not $null$}
        \State Infer action-related robotic capabilities
        \If{capability is mismatch}
            \State Abort execution
        \EndIf
        \State Infer how to call the action according to its type
        \State Parametrize the action call by querying the knowledge graph
        \State action = action.next
    \EndWhile
\EndProcedure
\end{algorithmic}
\end{algorithm}

\subsection{Inference by querying the knowledge graph}

As aforementioned, lines $8$ and $9$ of Algorithm~\ref{alg:mainloop} utilize SWI-Prolog queries to extract relevant knowledge from the knowledge graph. In this section, we detail the structure of these queries and explain how the corresponding action calls are generated after retrieving the necessary knowledge.

The process begins with the RCE checking for the next action to be executed in the action recipe. \textit{Action Recipes} consist of statements that define a sequence of actions, specifying the order in which tasks should be performed. To identify the next action, the system queries the action ordering to find the "preceding action" that corresponds to the action that has just been executed:

\begin{lstlisting}[style=prolog]
?-rdf(Order, knowrob:'occursBeforeInOrdering', OldTask), 
 rdf(Order, knowrob:'occursAfterInOrdering', Task).
\end{lstlisting}

Next, the RCE performs capability matching. It verifies whether the robot possesses the necessary capabilities to execute the identified action by querying the knowledge graph as follows:

\begin{lstlisting}[style=prolog]
?-rdf(Task, rdfs:subClassOf, Restriction), 
 rdf(Restriction, owl:onProperty, knowrob:'performedBy'), 
 rdf(Restriction, owl:hasValue, Robot), 
 rdf(Robot, srdl2:'hasCapability', Capability).
\end{lstlisting}

The required capabilities for a given action may include specific skills such as $srdl2:Navigate$ or $srdl2:Grasping$. If the robot possesses the required capabilities, the next step involves determining how to invoke the corresponding action. This is done through the following query:

\begin{lstlisting}[style=prolog]
?- rdf_reachable(Task, rdfs:subClassOf, ActionType), 
 rdf(ActionType, roboearth:'apiFunction', ApiFunction), 
 rdf(ActionType, roboearth:'apiParameter', ApiParameter).
\end{lstlisting}

Once the appropriate action is identified, the robot control executive infers additional contextual knowledge from the knowledge graph to parameterize the action call. For instance, the robot determines the location of the object relevant to the current task using the following Prolog query:

\begin{lstlisting}[style=prolog]
?-rdf(Task, rdfs:subClassOf, Restriction), 
 rdf(Restriction, owl:onProperty, knowrob:'objectActedOn'), 
 rdf(Restriction, owl:hasValue, Object), 
 rdf(Object, knowrob:'location', Loc).
\end{lstlisting}

Similarly, the robot queries the target location for the task, as illustrated in the following query:

\begin{lstlisting}[style=prolog]
?-rdf(Task, rdfs:subClassOf, Restriction), 
 rdf(Restriction, owl:onProperty, knowrob:'toLocation'), 
 rdf(Restriction, owl:hasValue, Furniture), 
 rdf(Furniture, rdf:type, knowrob:'Bed-PieceOfFurniture')
 rdf(Furniture, knowrob:'location', Loc).
\end{lstlisting}

\section{EXPERIMENTS}

\subsection{Experimental Framework}

\begin{figure}[!t]
    \centering
    \includegraphics[width=\linewidth]{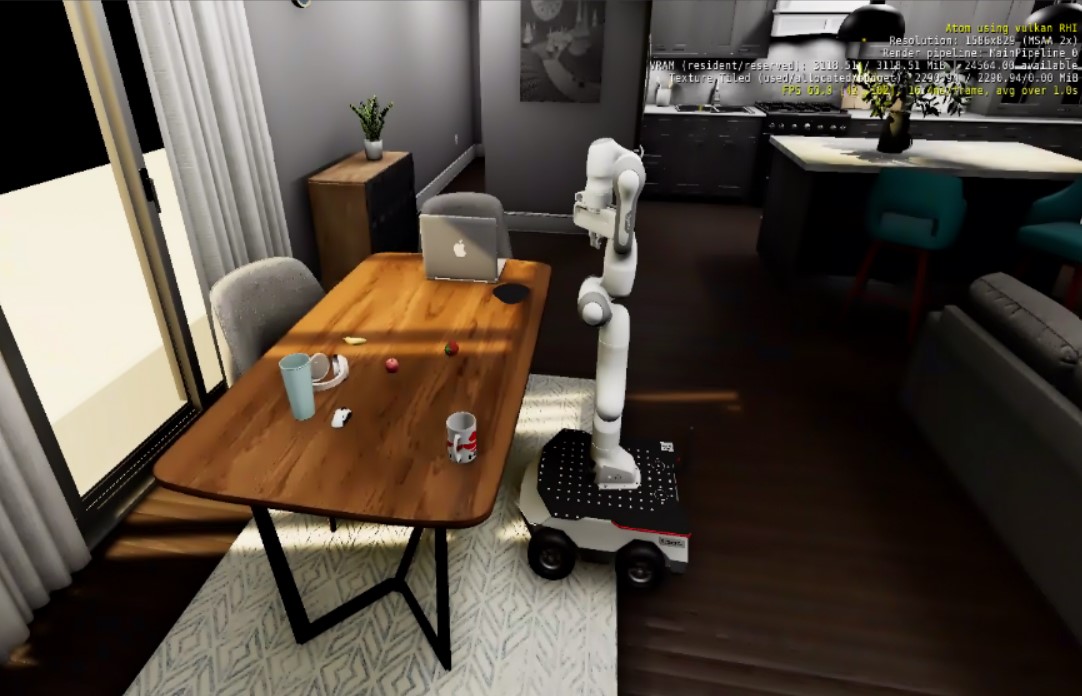}
    \caption{\small The household environment, Loft, and the robotic platform we are using in our experiments.}
    \label{fig:setup}
\end{figure}

We conducted experiments in a robotics simulation developed using the Open 3D Engine (O3DE)\footnote{https://o3de.org/}, as shown in Figure~\ref{fig:setup}. O3DE is an open-source, cross-platform game engine designed to create high-quality, real-time 3D experiences, including simulations, virtual worlds, and games. In addition to its advanced rendering, physics, and networking capabilities, O3DE provides a comprehensive robotics toolset. This includes the ROS2 Gem, which facilitates seamless integration with the Robot Operating System (ROS2), enabling the simulation, testing, and visualization of robotic systems in dynamic and interactive environments.

\begin{promptbox}
\textit{Can you provide a detailed \texttt{\{action\}} action recipe for a single robot within the Knowrob ontology as an OWL ontology encoded in RDF triples in an XML file, including logical partial strict orderings of subtasks? You need to generate such action recipes encoded in RDF triples in the XML file based on user prompts. Populate the action recipes with a detailed set of subtasks as possible. For generation, you must use OWL class names supplied in the embeddings. The objects related to the generated actions should also be included.}
\end{promptbox}

\subsection{Pick and Place}

As the first set of experiments, the user prompts the robot to perform pick-and-place tasks with vague descriptions. The expected outcome is that the system generates an \textit{action recipe} which accurately identifies the goal and establishes a valid high-level plan to achieve it.

\begin{figure*}[!t]
    \centering
    \includegraphics[width=\linewidth]{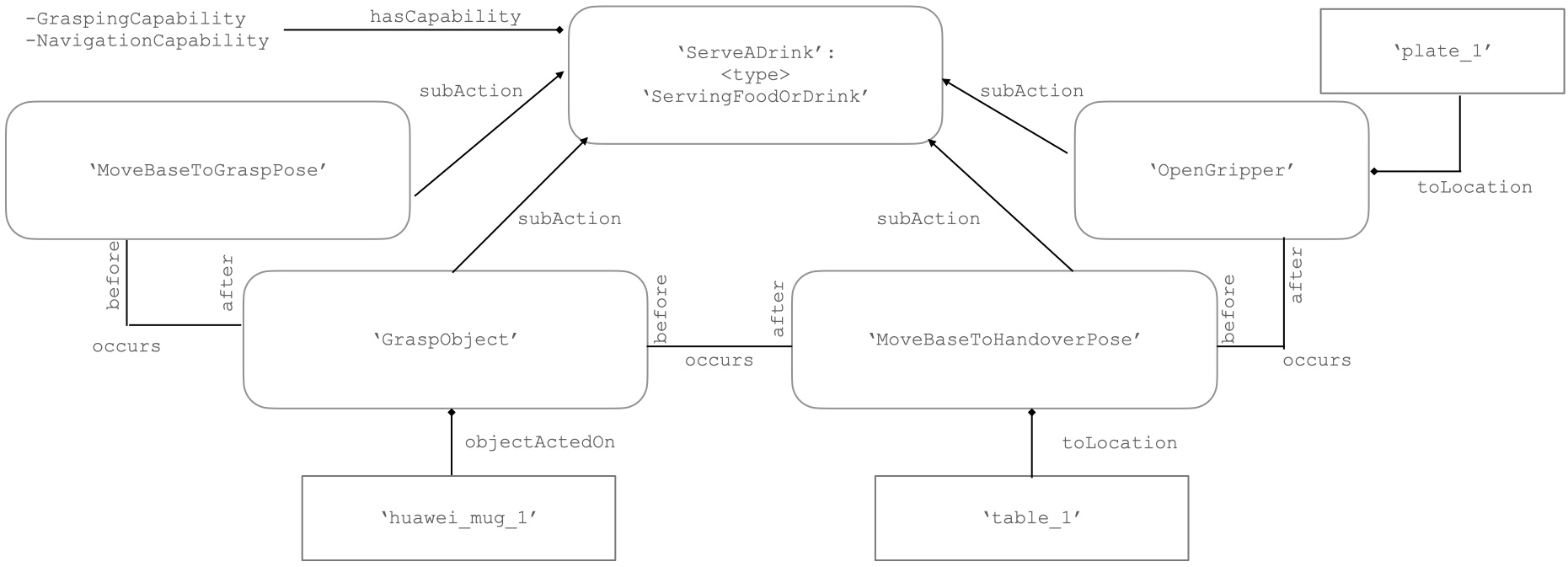}
    \caption{\small Graph representation of the \textit{action recipe} for \textit{“Serve me a drink”}, detailing required robot capabilities, subactions, and their temporal relationships (e.g., before-after). Key parameters, such as $objectActedOn$, are also specified for effective action parametrization.}
    \label{fig:recipe1}
\end{figure*}
\subsubsection{User Prompt: 'Serve me a drink'}
In this scenario, the user issues a simple prompt: "Serve me a drink", implying the retrieval of a liquid container referred to as a "drink." Since the term "drink" lacks a precise definition in the knowledge graph, the \textit{action recipe} must identify the specific object to be manipulated.

By referencing the digital twin knowledge, CodeLLaMa selects an appropriate object type from the available options and generates a pick-and-place \textit{action recipe} that specifies the correct object. In a scene containing various items, such as a vase, a strawberry, an apple, a Huawei mug, and headphones, the system identifies the most appropriate object—the Huawei mug. The key components of the generated \textit{action recipe} are as follows:

\scriptsize
\begin{verbatim}
<knowrob:DrinkingBottle 
  rdf:about="&roboearth;huawei_mug_1"/>
<knowrob:Plate 
  rdf:about="&roboearth;plate_1"/>
<knowrob:Bed-PieceOfFurniture 
  rdf:about="&roboearth;table_1"/>

<!-- Define subactions and ordering constraints -->
<owl:Class rdf:ID="ServeADrink">
  <rdfs:subClassOf 
    rdf:resource="&roboearth;ServingFoodOrDrink"/>
  <rdfs:label rdf:datatype="&xsd;string">
    serve a drink</rdfs:label>
  <owl:Class>
    <owl:intersectionOf rdf:parseType="Collection">
      <!-- Subaction constraints for ServeADrink -->
      <owl:Restriction>
       <owl:onProperty rdf:resource="&knowrob;subAction"/>
       <owl:someValuesFrom 
          rdf:resource="#MoveBaseToGraspPose"/>
      </owl:Restriction>
      <owl:Restriction>
       <owl:onProperty rdf:resource="&knowrob;subAction"/>
       <owl:someValuesFrom 
          rdf:resource="#GraspObject"/>
      </owl:Restriction>
      <owl:Restriction>
       <owl:onProperty rdf:resource="&knowrob;subAction"/>
       <owl:someValuesFrom 
          rdf:resource="#OpenGripper"/>
      </owl:Restriction>
      <owl:Restriction>
       <owl:onProperty rdf:resource="&knowrob;subAction"/>
       <owl:someValuesFrom 
          rdf:resource="#MoveBaseToHandoverPose"/>
      </owl:Restriction>
      <owl:Restriction>
       <owl:onProperty rdf:resource="&knowrob;subAction"/>
       <owl:someValuesFrom rdf:resource="#HandoverObject"/>
      </owl:Restriction>
    </owl:intersectionOf>
  </owl:Class>
</owl:Class>
\end{verbatim}

\normalsize

The more detailed graph representation is illustrated in Figure~\ref{fig:recipe1}. This LLM-generated \textit{action recipe} outlines the task of serving a drink by outlining the necessary subactions and constraints, such as grasping the correct object (in this case, the Huawei mug) and performing a series of operations like moving to the grasping position, picking up the object, and handing it over. These definitions directly inform the Prolog queries. For instance, the ordering of subactions can be determined by the $knowrob:$   $'occursBeforeInOrdering'$ and $knowrob$ $'occursAfterInOrdering'$ predicates, ensuring that the robot performs the steps in the correct sequence. Additionally, the Prolog queries about the robot's capabilities ($srdl2:$ $'hasCapability'$) can be answered by referencing the XML's description of which robot is responsible for the action and the subactions it performs. Similarly, the object involved in the action (the Huawei mug) and its location (on a table) are explicitly defined in the XML and can be retrieved through the Prolog query using $knowrob:$ $'objectActedOn'$ and $knowrob:$ $'toLocation'$. Finally, the query about target locations for the object can reference the furniture item (table) defined in the action recipe. This structured approach ensures that both the task planning and execution steps are grounded in the detailed information encoded in the \textit{action recipe}.

\begin{figure}[!t]
    \centering
    \includegraphics[width=\linewidth]{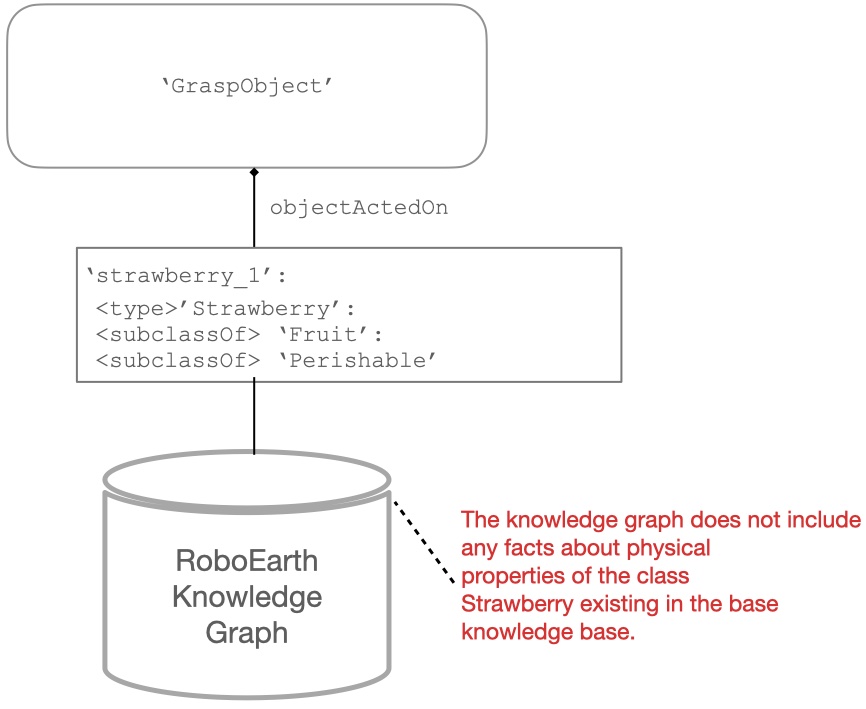}
    \caption{\small Target object for $GraspObject$ for \textit{Serve me the red small fruit} is $strawberry\_1$. As physical properties of fruits do not exist in the RKG, \textit{CodeLLaMa} makes the inference of description matching and explicitly states $strawberry\_1$ as the $objectActedOn$.}
    \label{fig:recipe2}
\end{figure}

\subsubsection{User Prompt: 'Serve me the red small fruit'}
In the second use case, the user prompts the system with the request, "serve me a red small fruit." Since attributes such as "red" and "small" are not explicitly defined in the knowledge graph, the system must infer the appropriate object based on contextual understanding.

By leveraging both the indexed digital twin knowledge and the available knowledge gained during training and fine-tuning, the environment specific CodeLLaMa identifies the strawberry as the most suitable object for the request. The generated \textit{action recipe} is similar to the one in Figure~\ref{fig:recipe1}, with the primary difference being in the $objectActedOn$ relation within the $GraspObject$ predicate. Despite the absence of such specific fruit descriptions in the base RKG, CodeLLaMa’s ability to generalize enables it to generate a pick-and-place \textit{action recipe} that selects the strawberry as the object to be manipulated.

The \textit{action recipe} for serving a red small fruit specifies the object of interest as $strawberry\_1$ and outlines a sequence of steps for the robot to complete the task. Initially, the recipe defines the robot's required capabilities, such as grasping, picking, and placing objects. The recipe then specifies a series of subactions, wherein the robot is tasked with picking up the strawberry, identified as the correct "red small fruit," and placing it on a table. Additionally, the recipe includes temporal ordering constraints to ensure that the robot follows the correct sequence of actions: first grasping the strawberry, then placing it at the desired location.


\subsection{Removing Objects from the Dining Table}
In these experiments, the robot is tasked with cleaning the objects on top of the dining table. To achieve this, we employ a two-step approach. First, the user prompts the system with the command, "perceive objects on the dining table." Upon execution of the corresponding \textit{action recipe}, the digital twin knowledge within the RKG is updated accordingly. Second, the user prompts the robot to remove the perceived objects from the table.

\subsubsection{User Prompt: 'Perceive Objects on the Dining Table}
\begin{figure}[!t]
    \centering
    \includegraphics[width=\linewidth]{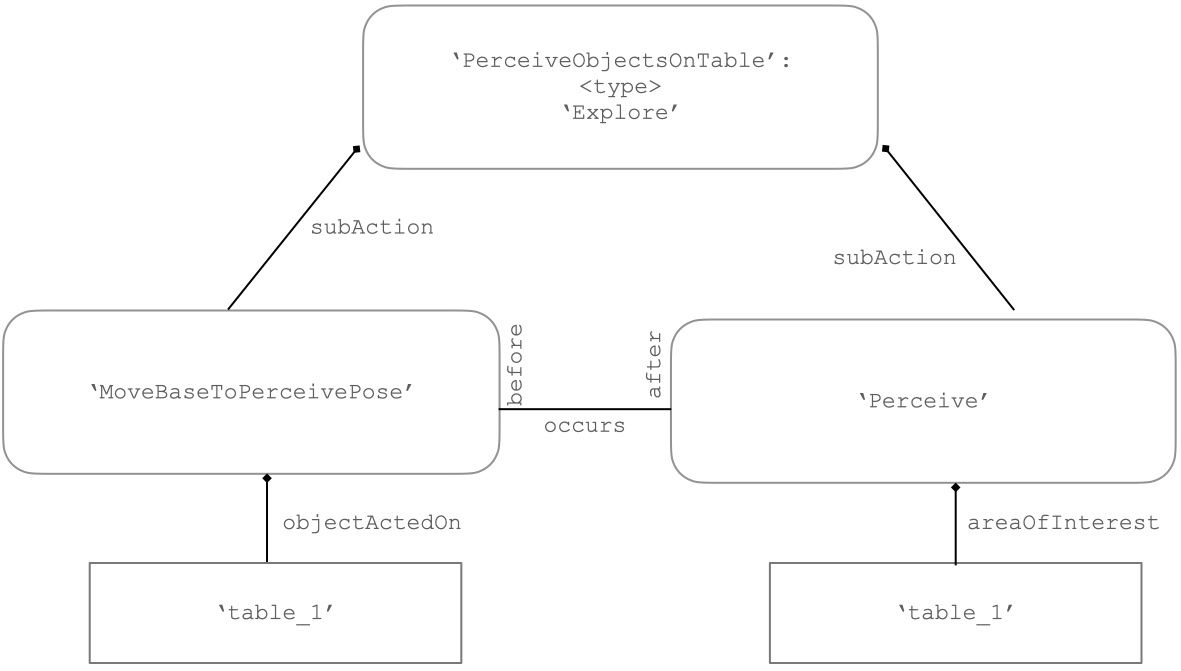}
    \caption{\small Graph representation of the \textit{action recipe} for \textit{“Perceive Objects on the Dining Table”}.}
    \label{fig:recipe3}
\end{figure}

This perception task results in a relatively straightforward \textit{action recipe}, which includes the robot navigating to the dining table and executing the perception routine (Figure~\ref{fig:recipe3}).

Following the execution of this routine, the perception results are registered into the RKG, updating the digital twin knowledge through Prolog queries. An example of such a query is as follows:
 \begin{lstlisting}[style=prolog]
?-rdf_assert(P, rdf:type, knowrob:'Perception'), 
 rdf_assert(P, knowrob:'location', Loc), 
 rdf_assert(P, knowrob:'occursAt', T), 
 rdf_assert(P, knowrob:'objectActedOn', roboearth:'huawei_mug_1').
\end{lstlisting}

In this assertion query, $Loc$ and $T$ represent the location of the object and the time of perception, respectively, both of which are bound (or initialized) by the perception routine.

\subsubsection{User Prompt: 'Clean the Dining Table'}
\begin{figure*}[!t]
    \centering
    \includegraphics[width=\linewidth]{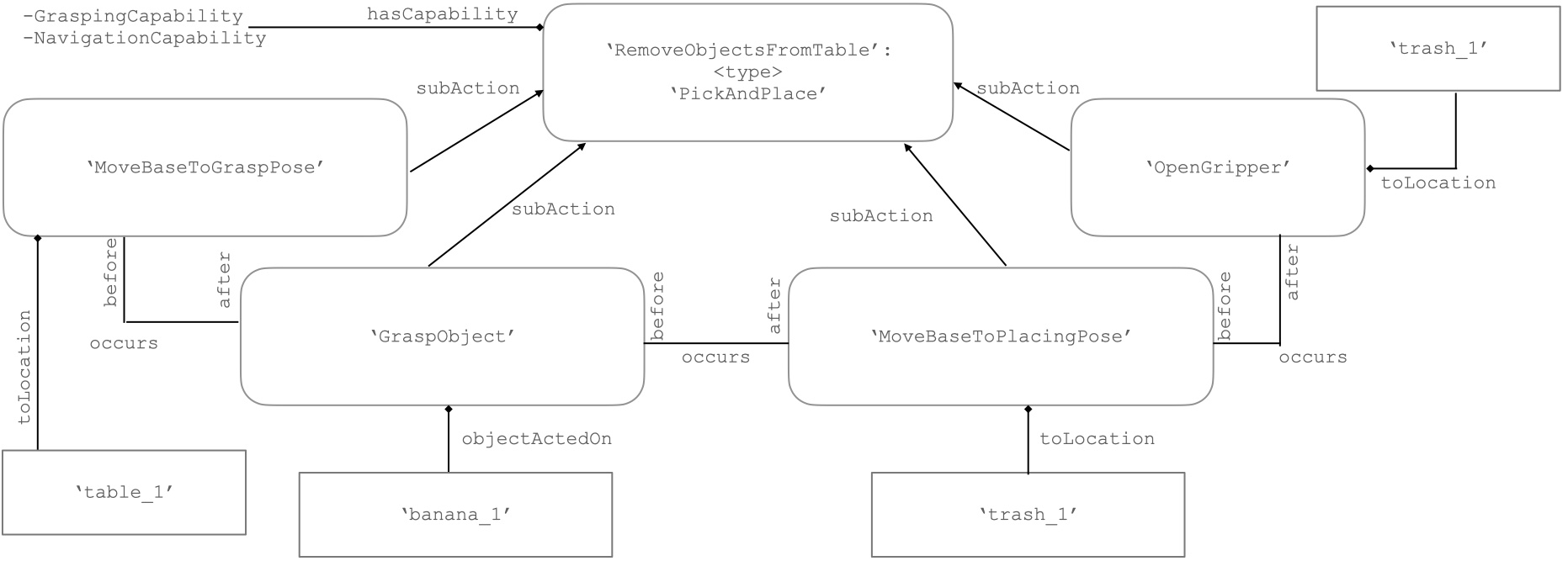}
    \caption{\small Graph representation of one pick-and-place action sequence in the \textit{action recipe} for \textit{“Clean the Dining Table”}.}
    \label{fig:recipe4}
\end{figure*}

Once the objects on the table have been perceived, the robot is tasked with "cleaning the dining table" using the most up-to-date digital twin knowledge. The resulting \textit{action recipe} consists of a series of pick-and-place actions for each item on the table. In this case, the designated target destination for all objects is the $trash\_1$ (Figure~\ref{fig:recipe4}).

\subsection{Hallucinations}
The above experiments include the \textit{action recipes} which are useable and aligned with the RKG. However, this is not always the case as one of the primary challenges associated with the LLMs is the occurrence of hallucinations. In the proposed pipeline, these hallucinations are more readily detectable, as the system operates within a semantic layer. The most common types of hallucinations are: (1) the resulting \textit{action recipe} contains an object assertion (ABOX) that is not represented in the digital twin knowledge, and (2) the \textit{action recipe} includes a terminology (TBOX) that does not exist within the RKG. In such cases, Prolog queries typically return a failure or no result, prompting the system to flag the corresponding \textit{action recipe} as unusable and restart the pipeline by re-prompting the system. Table 1 presents the number of usable action recipes for each prompt after 100 iterations.

\begin{table}[h]
\centering
\begin{tabular}{|c|c|c|}
\hline
\textbf{} & \textbf{No of trials} & \textbf{Useable Recipes}  \\ \hline
\textbf{Serve me a drink}       & 100   & 88  \\ \hline
\textbf{Serve me the red small fruit}       & 100   & 76   \\ \hline
\textbf{Perceive Objects on the D. Table}       & 100   & 97 \\ \hline
\textbf{Clean the Dining Table}       & 100   &  73  \\ \hline
\end{tabular}
\label{fig:table}
\caption{Number of useable \textit{action recipes} after prompting the required action 100 times.}
\end{table}

\subsection{Evaluation}

To evaluate the scalability of our approach, we compare our results with SMART-LLM~\cite{kannan2024smartllmsmartmultiagentrobot}, one of the most recent high-level planners leveraging LLMs. For this comparison, we selected SMART-LLM with the Llama2-70b model, as CodeLLaMa is also based on the Llama2 architecture. In contrast, our model utilizes the smaller Llama2-13b variant, allowing for a direct comparison across different model sizes.

As experimented in \cite{kannan2024smartllmsmartmultiagentrobot}, we use four task sets—\textit{elemental}, \textit{simple}, \textit{compound}, and \textit{complex} task sets from the AI-THOR2~\cite{kolve2022ai2thorinteractive3denvironment}. We sample 1,000 tasks from these sets and validate their usability formally by querying them through Prolog. These task descriptions are more concrete than the abstract prompts tested previously. In Table 2, we present a comparison of \textit{RecipeMasterLLM} and SMART-LLM (Llama2-70b) based on success rates (SR).

\begin{table}[h]
\centering
\begin{tabular}{|c|c|c|}
\hline
\textbf{} & \textbf{Ours} & \textbf{SMART-LLM  (Llama2-70b)}  \\ \hline
\textbf{Elemantal}       & 0.940   & 1.0  \\ \hline
\textbf{Simple}       & 0.871   & 0.75   \\ \hline
\textbf{Compound}       & 0.745   & 0.64 \\ \hline
\textbf{Complex}       & 0.781   &  0.63  \\ \hline
\end{tabular}
\label{fig:table2}
\caption{: Evaluation of the success rates for our \textit{RecipeMasterLLM} and the SmartLLM baseline for different sets of tasks.}
\end{table}

As shown in Table 2, our model outperforms SMART-LLM in all tasks except for the \textit{elemental} tasks, where hallucinations arise due to the smaller model size. In the remaining task sets, our model not only surpasses Llama2-based SMART-LLM but also other variants of SMART-LLM according to the results provided in \cite{kannan2024smartllmsmartmultiagentrobot}.

\section{CONCLUSIONS}
In this paper, we presented a novel approach for automating action description generation for robots, integrating fine-tuned LLMs with the RoboEarth Knowledge Graph (RKG) and Retrieval-Augmented Generation (RAG) techniques. This enables the creation of actionable, context-aware \textit{action recipes} that robots can use to autonomously plan and execute tasks. By combining LLMs with a standardized knowledge graph, we offer a scalable solution to the challenges of generating action plans in dynamic environments. Our experiments demonstrate the system's ability to handle diverse user prompts and generate grounded action plans, though challenges like hallucinations remain, which we address through automatic validation. Future work will aim to refine the system's robustness and extend its capabilities to more complex domains.

\bibliographystyle{IEEEtran}  
\bibliography{references}     

\end{document}